\documentclass[10pt,twocolumn,letterpaper]{article}

\usepackage{array}
\newcolumntype{P}[1]{>{\centering\arraybackslash}p{#1}}
\newcolumntype{M}[1]{>{\centering\arraybackslash}m{#1}}
\usepackage{graphicx}
\usepackage{booktabs}
\usepackage{array}
\usepackage[table,xcdraw]{xcolor}
\usepackage{amsmath,amssymb,bm} 
\usepackage{color}
\usepackage{xcolor}
\DeclareMathOperator*{\argmax}{arg\,max}

\usepackage{wacv}
\usepackage{times}
\usepackage{epsfig}
\usepackage{graphicx}
\usepackage{booktabs}
\usepackage{amsmath}
\usepackage{amssymb}
\usepackage{nccmath}
\usepackage{mathtools}
\usepackage{enumitem}
\pdfoutput=1

\wacvfinalcopy 

\ifwacvfinal
\def\assignedStartPage{9876} 
\fi

\ifwacvfinal
\usepackage[breaklinks=true,bookmarks=false]{hyperref}
\else
\usepackage[pagebackref=true,breaklinks=true,colorlinks,bookmarks=false]{hyperref}
\fi

\ifwacvfinal
\else
\fi

\begin{document}

\title{SubICap: Towards Subword-informed Image Captioning}

\author{Naeha Sharif,
	Mohammed Bennamoun,
	Wei Liu\\
The University of Western Australia\\
Perth, Western Australia\\
{\tt\small naeha.sharif@research.uwa.edu.au}
\and
Syed Afaq Ali Shah\\
Murdoch University\\
Perth, Western Australia\\
{\tt\small Afaq.Shah@murdoch.edu.au}
}

\maketitle

\begin{abstract}
 Existing Image Captioning (IC) systems model words as atomic units in captions and are unable to exploit the structural information in the words. This makes representation of rare words very difficult and out-of-vocabulary words impossible. Moreover, to avoid computational complexity, existing IC models operate over a modest sized vocabulary of frequent words, such that the identity of rare words is lost. In this work we address this common limitation of IC systems in dealing with rare words in the corpora. We decompose words into smaller constituent units \textit{`subwords'} and represent captions as a sequence of subwords instead of words. This helps represent all words in the corpora using a significantly lower subword vocabulary, leading to better parameter learning. Using subword language modeling, our captioning system improves various metric scores, with a training vocabulary size approximately 90\% less than the baseline and various state-of-the-art word-level models. Our quantitative and qualitative results and analysis signify the efficacy of our proposed approach.	
\end{abstract}

\section{Introduction}
Image captioning is a challenging task that bridges two different domains of Artificial Intelligence (AI), Computer Vision (CV) and Natural Language Processing (NLP). It takes both visual as well as linguistic understanding for a system to translate visual information into well-formed sentences. Over the past decade, numerous frameworks have been proposed for captioning \cite{anderson2017bottom}, \cite{mao2014deep}, \cite{yao2016boosting}, amongst which encoder-decoder based neural models have been very popular. In this particular framework, encoder transforms the visual input into a visual embedding, whereas, the decoder uses the encoded embedding as an input to generate text. 

\begin{figure}[t]
	\centering
	\includegraphics[width=8cm]{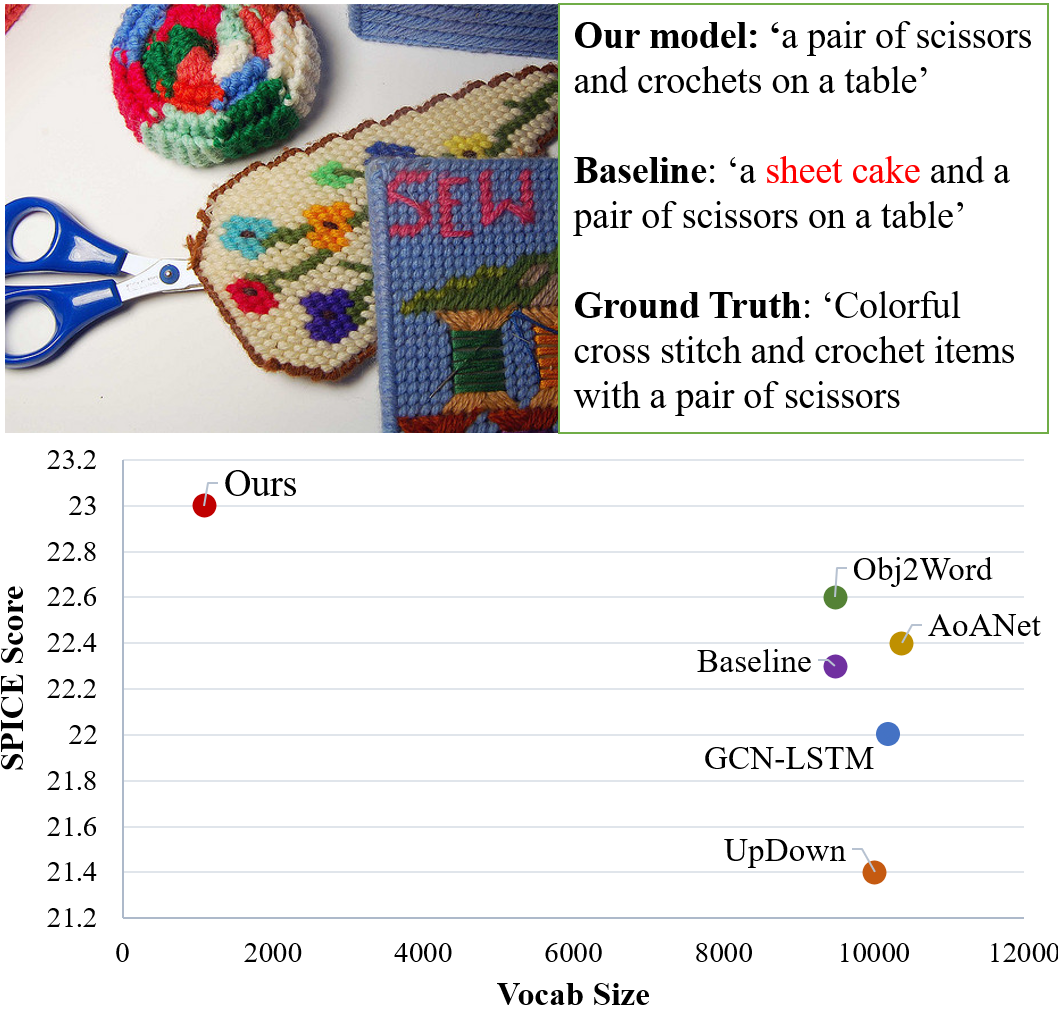}
	\caption{Our proposed model, which uses subword information, generates better captions compared to the baseline, which operates at the word-level. This example shows captions generated by our model and the baseline for an image. Our model generates the rare word `crochets' in the output caption. Compared to state-of-the-art models (UpDown \cite{anderson2017bottom}, GCN-LSTM \cite{yao2018exploring}, Obj2Word \cite{herdade2019image}, AoANet \cite{huang2019attention}), our model achieves a higher SPICE score on the MSCOCO dataset, using a significantly smaller training vocabulary size.}
	\vspace{-1.5em}
	\label{fig:subword}
\end{figure}

Various encoder-decoder based models differ in the way  of encoding the visual information. While some have used features from penultimate layers of CNN classifiers \cite{mao2014deep}, others have focused on extracting the most relevant visual features, using attention \cite{anderson2017bottom}. Leveraging semantic concepts \cite{wu2016value}, object attributes \cite{yao2016boosting} and relationships \cite{yao2018exploring} has also garnered interest. In short, a lot of work has been done in terms of representing the visual input into meaningful latent representation(s). Such representations play an important role in IC and can be reflective of a model's visual understanding.

The decoder in an encoder-decoder based IC framework is usually a language model, which learns to generate captions as a sequence of words/tokens, given a visual encoding. Existing State-of-The-Art (SOTA) captioning systems have modeled individual words as atomic units. Captions have been treated as a sequence of words, completely ignoring the fact that words are also composed of smaller units called \textit{morphemes}. To deeply understand the meaning of a caption, it can be useful to also leverage upon the composition of words. Limitations of the existing word-level captioning models are: 1) they assume a limited vocabulary of frequent words, thus poorly generalize rare words into a single category of ‘unknown words’, 2) these models are unable to exploit the relationship between the composition and meaning of words. For example the word `dog' and `dogs' might be considered as two individual words by a word-based model. However, `dogs' is actually a plural of `dog' and is composed of `dog', and `s'. 

To alleviate these major limitations of the existing captioning models, we introduce subword-informed image captioning (SubICap). Subword units have become prevalent in various NLP models over the past few years \cite{sennrich2015neural}, \cite{ zhu2019importance}.  The subword-informed models represent words as a composition of \textit{subwords}. Identifying similar subwords amongst words, can be helpful in exploiting the relationship between word segments, their meanings and compositionality. Since, the number of plausible subwords in a corpora is comparatively minuscule vs. the number of plausible words, subword-level models have a smaller number of trainable parameters compared to the word-level models. Moreover, subword-level models can effectively represent rare words. Since words that have similar substrings (such as `flower' and `flowers') can share similar representations, the model can see words that share the same sub-strings with rare words during training, resulting in rare words having more informative representations. In addition, the average frequency of subwords is higher than the frequency of words, which helps reduce the sparsity issue in the corpora and leads to better parameter learning and probability estimates. Our contributions in this work are: 

\vspace{-0.5pc}
\begin{enumerate}[noitemsep]
\item We propose subword-based language modeling for image captioning as a better alternative to word-level language modeling, to adequately handle rare words in the training set, reduce the training parameters (Sec.~\ref{subsection:Influence of the Vocabulary Size}) and to improve the quality of captions (Sec.~\ref{subsection:Caption Uniqueness and Descriptiveness},~\ref{subsection:Qualitative Analysis}).\useshortskip
\item To the best of our knowledge, this is the first work that employs a learned subword tokenizer for captioning.
\item We demonstrate the impact of choosing different vocabulary sizes for the subword tokenizer on various metric scores, number of learning parameters and percentage of unique captions (Sec.~\ref{subsection:Influence of the Vocabulary Size}). 
\item An in-depth quantitative (Sec.~\ref{subsection:Caption Uniqueness and Descriptiveness},~\ref{subsection:Quantitative Analysis}) and qualitative analysis (Sec.~\ref{subsection:Qualitative Analysis}) to show that our model improves upon the word-level baseline not only in terms of popular metric scores but also on uniqueness and descriptiveness of captions.
\end{enumerate}

\section{Related Literature}

\subsection{Image Captioning}
Image captioning is a multi-modal task that aims at translating visual information into linguistic descriptions. Over the past decade, image captioning has seen remarkable progress. Image Captioning (IC) leverages the progress made in both CV and NLP, by using SOTA deep visual \cite{ren2015faster}, \cite{khan2018guide} and linguistic models \cite{greff2016lstm} as building blocks of the IC framework. Moreover, various strong IC frameworks, such as encoder-decoder \cite{karpathy2015deep}, \cite{vinyals2015show} and attention-based models \cite{anderson2017bottom}, \cite{huang2019attention}, \cite{xu2015show} have been inspired from a closely related domain, i.e., Machine Translation (MT). Various language models such as RNNs \cite{mikolov2010recurrent} and Transformers \cite{vaswani2017attention} that have shown promise in MT, have worked very well for IC as well. 

\subsection{Understanding Composition of Words}
The task of image captioning involves both visual and linguistic understanding. Our focus is more towards improving the later, so that the model can better express the visual information.  Linguistic understanding in humans encompasses various aspects. Though it is beyond the scope of this research to cover all of those,  we want to shed light on one of the stepping stones in language development i.e., acquiring phonemic awareness. Phonemic awareness involves the understanding that words are composed of sounds, sounds are made up of letter(s) (subwords), and letters fit together in different ways to convey different meanings \cite{tankersley2003threads}. The ability to decompose words or put them back together to make new words is one of the foundational skills that helps in acquiring meaning from text \cite{tankersley2003threads}.  Therefore, training language models to compose words from subwords can in some way be considered analogous to teaching humans to compose words out of phonemes. Moreover, a language model that learns to compose sentences using subword tokens, develops the understanding of both sentence as well as word composition. 

While, phoneme is the smallest units of sound, morpheme is the smallest meaningful part of a word \cite{vania2020understanding}. In order to understand the meaning of a caption, it is important to understand the meaning and composition of individual words. However, supervised morphological analysis is very expensive and not that straightforward \cite{vania2020understanding}. Identifying similar subwords which are \textbf{‘characters, n-grams, or word segments'}, can be helpful in exploiting the relationship between word segments and their composition. Zhu. et al, \cite{ zhu2019importance} showed that subword-informed models are useful across all language types, with better performance over subword-agnostic word embeddings.

\subsection{Subword Segmentation}
Since it is very challenging to perform supervised morphological segmentation of words, unsupervised methods to autonomously discover segmentations for the words are widely used \cite{sennrich2015neural}. Subwords obtained from such methods often resemble linguistic morphemes \cite{creutz2006induction}. FastText \cite{bojanowski2017enriching} is a \textbf{character n-gram} model, which represents each word as a sum of character $n$-grams ($n$ belongs to \{3,4,5,6\}). \textbf{Character-level} models learn to form words from characters instead of character strings/segments.

\begin{figure}[t]
	\centering
	\includegraphics[width=8.25cm]{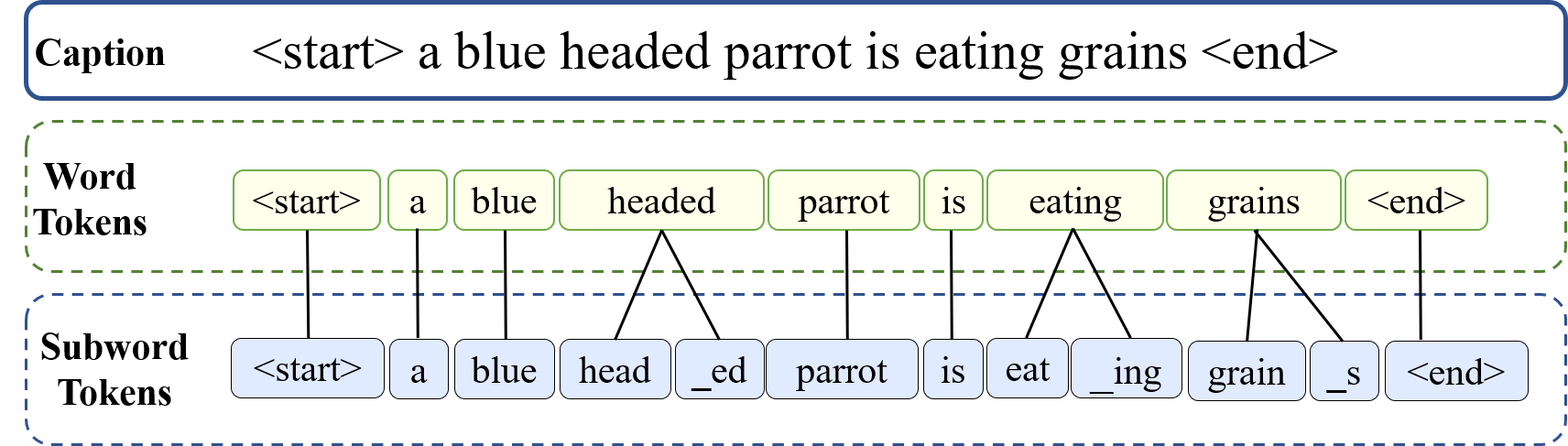}
	\caption{Word-level and subword-level tokenization of a caption}
	\vspace{-1.5em}
	\label{fig:subword}
\end{figure}
Subword models that represent words as segments outperform character-level models \cite{mikolov2012subword}, have zero out-of vocabulary rate, and are smaller in size. 
Byte Pair Encoding (BPE) \cite{sennrich2015neural} is one of the most popular subword tokenization technique, which given a dataset, learns a fixed-size vocabulary of subwords. Initially, it splits the sentences into individual characters, then it iteratively merges the pairs of characters based on their frequency of occurrence; resulting in more frequent words not being segmented. However, BPE is deterministic and splits words into unique sequences, which may prevent a model from learning the better composition of words. A similar word segmentation algorithm which is used in BERT \cite{devlin2018bert} is called WordPiece \cite{schuster2012japanese}. While both BPE and WordPiece require data to learn word segmentations,  WordPiece forms the new subwords based on likelihood instead of pairing frequency. 

Unigram language model \cite{kudo2018subword} is a more flexible segmentation algorithm compared to BPE. It is based on a probabilistic language model and is capable of outputting multiple subword segmentations along with their probabilities. The final vocabulary contains all individual characters in the corpus and a mixture of subwords and words. In this paper, we use unigram language model for subword segmentation. We also provide a comparison between different subword segmentation methods in the supplementary material.

\subsection{Leveraging Subword Information}
Subword-level information is useful in various NLP tasks such as learning word representations \cite{ zhu2019systematic}, sequence tagging \cite{devlin2018bert} and machine translation \cite{sennrich2015neural}. 
The subword-level architectures leverage the structural knowledge of words, assuming that a word’s meaning can be inferred from the meaning of its constituents (i.e., subwords). Subword-based neural architectures decrease data sparsity by relying on parameterization at the subword-level \cite{kudo2018subword}. 

To the best of our knowledge, the existing encoder-decoder based IC approaches for English language ignore the syntactic composition of words from subwords/morphemes, except for \cite{tan2019comic}. Assigning only a single vector to each word causes the data sparsity problem. Subword-agnostic word-level representation models do not take these structural features into account and are unable to represent rare words accurately, or unseen words at all. The closest research to our work is \cite{tan2019comic}, which uses radix-encoding to transform words into a higher-base to train a compact LSTM-based IC framework. In contrast to \cite{tan2019comic} we use a `learned subword segmentation' algorithm along with a `transformer-based' IC baseline \cite{herdade2019image}.

\subsection{Language Models}
Given a visual representation, an IC model seeks to translate it into a sequence of words. For that task, various language models (LMs) have been used in the literature such as, Maximum-Entropy LM \cite{fang2015captions}, Recurrent Neural Networks (RNNs) \cite{vinyals2015show} and LSTMs \cite{anderson2017bottom}. Though LSTMs/RNNs have been a popular choice in IC, they struggle in handling long-term dependencies and are slow to train. Recently, Transformer networks \cite{vaswani2017attention} have been used in IC \cite{sharma2018conceptual}, \cite{herdade2019image}, which address various limitations of LSTMs and have also shown promising results in MT \cite{vaswani2017attention}, text generation \cite{devlin2018bert}  and understanding \cite{radford2019language}. Transformers are becoming the de facto choice for sequence modeling in NLP. Therefore, we also use a Transformer based IC model \cite{herdade2019image}, as a baseline for our work.

\begin{figure*}[t]
	\centering
	\includegraphics[width=0.9\textwidth]{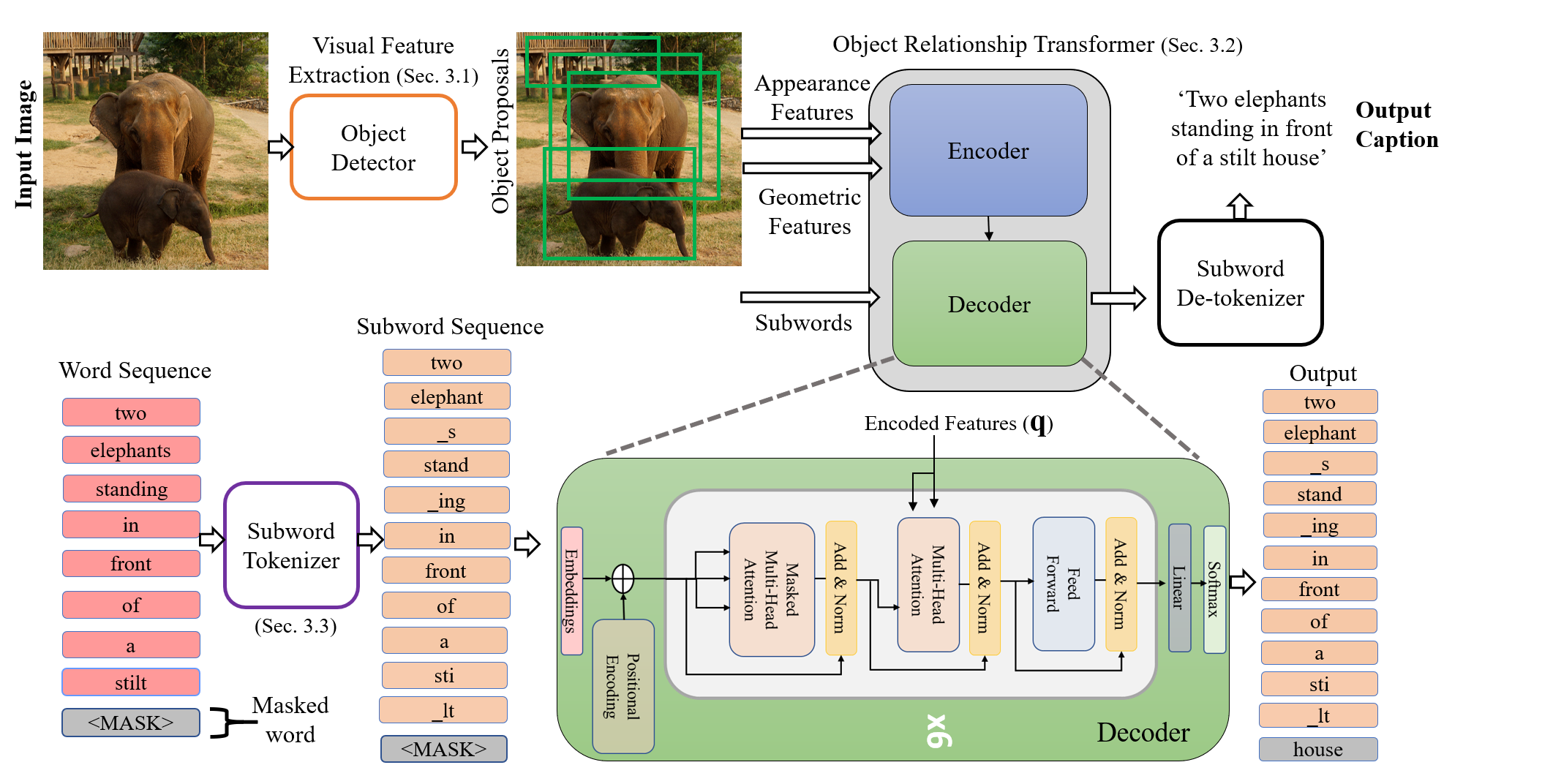}
	\vspace{-0.5em}
	\caption{Overview of our proposed model, which consists of an encoder-decoder based Transformer pipeline \cite{herdade2019image}. During the training process, the decoder leverages encoded embedding $\textbf{q}$ and a sequence of subwords to predict the masked subword. The subwords generated by the decoder are detokenized to obtain the output caption. For deeper insight into the encoder architecture, we refer readers to \cite{herdade2019image} }
	\vspace{-1em}
	\label{fig:Architecture}
\end{figure*}

\section{Subword-informed Image Captioning}

Our proposed approach involves using subword-information for IC. Our model represents each caption as a sequence of subwords, where subwords are a combination of strings such as `ing' or `ed'. Figure~\ref{fig:subword} shows an example of word-level, and subword-level tokenization of a caption. Tokenization refers to the process of segmenting a stream of characters into individual units, known as tokens. Tokenization is one of the most important steps in language modeling because it impacts the way a model sees the textual input, i.e., as a sequence of words, subwords or characters. In this work, we transform an existing word-level IC model \cite{herdade2019image} into a subword-level model. Figure~\ref{fig:Architecture} shows the architecture of our model.

\subsection{Visual-Geometric Feature Extraction}

Given an input image \textit{I}, a number of bounding boxes are generated for the detected objects. A subset of these overlapping  bounding boxes are discarded if their intersection-over-union (IoU) exceeds a threshold \textit{t} (0.7 in our case).  Furthermore, we only keep those bounding boxes which have a class prediction probability greater than a threshold of 0.2. For each bounding box we extract a 2048-dimensional feature vector by applying mean-pooling over the spatial dimension. The extracted feature vectors are finally processed through an embedding layer to generate visual appearance embeddings. These visual appearance embeddings are used as input tokens to the first Encoder layer of the Transformer. For each bounding box, we also extract geometric features, such as center coordinates ($x_l, y_l$), ($x_m, y_m$)), widths ($w_l, w_m$) and heights ($h_l, h_m$). 

\subsection{Object Relational Transformer}

Our transformer based architecture, consists of an encoder and a decoder, both of which are comprised of stacked multihead self-attention and point wise, fully connected feed forward layers. The encoder transforms sequence of visual appearance \{$y_1, y_2,..., y_N$\} and geometric \{$g_1, g_2,..., g_N$\} embeddings to a sequence of continuous embeddings $\textbf{q}$ $=$ ($q_1$, $q_2$, .., $q_N$). Given \textbf{q}, the decoder generates a sequence of tokens ($t_1$, $t_2$, .., $t_s$), using output tokens (previously generated and masked) as an additional input \cite{vaswani2017attention}. In contrast with \cite{herdade2019image}, the decoder in our model generates subwords, instead of words as output tokens.

The encoder consists of six layers, where each layer is composed of a multi-head self-attention layer and a feed-forward neural network. The multi-head self-attention layer consists of eight identical heads, each of which computes a query Q, key K and value V for the $N$ input token embeddings, given by:\useshortskip
\begin{align}
Q = Y{W_Q}, K = Y{W_K},   K = Y{W_V} 
\end{align}
\noindent where, Y is the input token matrix containing the visual embeddings \{$y_1,y_2,.., y_N$\} and ${W_Q}$, ${W_V}$, and ${W_K}$ are the learned projections. The attention weight matrix $\theta$ for the visual features is formulated as:
\begin{align}
\theta_{A}=\frac{QK^T}{\sqrt{d_{k}}}
\end{align}
\noindent where, $\theta_{A}$ is an attention weight matrix ($N\times N$), whose element  $\omega_{A}^{lm}$ corresponds to the attention weights between the $l^{th}$ and $m^{th}$ tokens. $\theta_{A}$ is modified by incorporating relative geometric features \{$g_1, g_2,..., g_N$\} of objects. The visual appearance-based attention weights $\omega_{A}^{lm}$  between the $l^{th}$ and $m^{th}$ object are multiplied by geometric attention weights, given by:
\begin{align}
\omega_{G}^{lm} = ReLU (Emb(\lambda(l,m) )W_G)
\end{align}

\noindent where, $Emb(\cdot)$ computes a positional embedding, described in \cite{herdade2019image}, $W_G$ is a transformation matrix and  $\lambda(l,m)$ is a displacement vector corresponding to objects \textit{l} and \textit{m} (See supplementary material for details). The geometric and visual-appearance attention weights $\omega_{G}^{lm}$ and $\omega_{A}^{lm}$ respectively are combined to form visual-geometric attention wights $\omega^{lm}$, which are computed as:

\begin{align}
\omega^{lm} = \frac{\omega_{G}^{lm} \exp (\omega_{A}^{lm})} {\sum_{i=1}^{N}\omega_{G}^{li} \exp(\omega_{A}^{li})}
\end{align}

The output of each attention head is formulated as:
\begin{align}
Head(Y)=softmax(\theta)V
\vspace{-1.3em}
\end{align}

\noindent where, $\theta$ is an NxN matrix, whose elements are the visual-geometric attention wights $\omega^{lm}$. The outputs of all the attention heads (8 in our case) are concatenated and then multiplied to a learned projection matrix $W_o$.
Next, the output of the self-attention layer (MultiHead) is fed to a point-wise feed-forward network (FFN). The FFN consists of two linear projection layers with a ReLU activation function in between.

The decoder uses the encoded visual-geometric embeddings $\textbf{q}$ generated from the last encoder layer, to generate a sequence of subwords. We refer the reader to \cite{vaswani2017attention} for more details on the decoder, as we use adopt their decoder architecture in this work.

\subsection{Subword Tokenization}
\label{subsection:Subword Tokenization}
In this work we use Unigram Language Model (ULM) \cite{kudo2018subword} to segment/tokenize words into a vocabulary of subwords. Given a set of captions, ULM models the probability of a subword sequence \begin{math}\textbf{s} = (s_1, s_2, ..., s_T) \end{math} as a product of the subword occurrence probabilities \begin{math} p(s_i)\end{math}. \useshortskip

\begin{align}
P(\textbf{s}) =\prod_{t=1}^{T}p(s_i)
\vspace{-1.3em}
\end{align}

\begin{align}
s^* = & \argmax_{s\in Z(S)} P(\textbf{s})
	\vspace{-0.5em}
\end{align}

\noindent where, $s^*$ is the most probable segmentation of the input sequence and is obtained via viterbi algorithm \cite{viterbi1967error}. $Z(S)$ is a set of subword candidates for the input sequence $S$.  In order to obtain the probability of subwords \begin{math} p(s_i)\end{math}, an Expectation-Maximization algorithm is used, and we refer the readers to \cite{kudo2018subword} for more details. 

For ULM-based segmentation, a desirable vocabulary size $k$ has to be pre-defined and then the model seeks to create a Vocabulary $V_u$ of subwords by following an iterative algorithm, which is discussed in \cite{kudo2018subword}. The final vocabulary $V_u$ obtained through the model contains a mixture of characters, subwords and words in the corpus. Using subword tokenization, we manage to re-represent the entire corpus vocabulary $V$ of size $j$ to a vocabulary $V_u$  of size $k$, where $k$ is significantly smaller than $j$. This reduction in vocabulary size also brings down the model complexity. For example, the training vocabulary used by the SOTA models \cite{anderson2017bottom}, \cite{herdade2019image} for the MSCOCO dataset is usually around 9,000-10,000 words, which we choose to represent in approximately 1000 subwords. This accounts to about 90\% reduction in the vocabulary size. See Table~\ref{tab:comparison} for results.

\textbf{Detokenization:} To allow for a deterministic recovery of words for output captions, a `\_' marker is attached at the start of subwords to represent the intra-word boundary. Subwords are concatenated to form complete words. For example in Fig.~\ref{fig:subword}, \textit{\_ed} will be combined with \textit{head} to form a word \textit{headed}. A sequence of words are then concatenated with whitespaces in between to form a caption.  

\subsection{Subword Language Modeling}
Given an image I and caption S, let  $\textbf{s}$ $=$ ($s_1$, $s_2$, ...,$ s_T$) be the segmented subword sequence, and $\textbf{q}$ $=$ ($q_1$, $q_2$, ..., $q_N$) be the visual-geometric embeddings corresponding to S and I respectively. Our captioning system models the caption generation probability as $P(S|I)$ = $P($\textbf{s}$|$\textbf{q}$)$, and generates target subword $s_t$ conditioned on the target history $s_{<T}$ and visual embedding:\useshortskip
\begin{align}
P(\textbf{s}|\textbf{q};\Theta) =\prod_{t=1}^{T} s_{i} P(s_t|\textbf{q},s_{<T};\Theta)
\vspace{-1.5em}
\end{align}
\useshortskip
\noindent where $\Theta$ is a set of model parameters. 
To generate a caption at the inference stage, we use beam search and apply post-processing (de-tokenization) to the output tokens. The post-processing involves converting the segmented subword sequence to a sequence of words.

\section{Implementation Details}

\subsection{Captioning Model}
 In this work, we use the PyTorch implementation of \cite{herdade2019image} as our baseline, which is a word-level IC model. For our model, we modify the baseline implementation, to include  subword modeling. All the models in our experiments are first trained for 30 epochs with a cross-entropy loss, with an initial learning rate of 5x$10^{-4}$, using ADAM optimizer, and a batch size of 15. We decay the initial learning rate by a factor of 0.8 every 3 epochs. We further train our models for 10 epochs, with a batch size of 10, using reinforcement learning \cite{rennie2017self} targeted at optimizing the CIDEr-D \cite{vedantam2015cider} score. We perform early stopping based on the best performance on the validation set. We ran our experiments on single NVIDIA Titan Xp GPU, on which it took about 1 day and 3.5 days for cross-entropy and reinforcement learning based training, respectively.

\subsection{Tokenizer}
 For our baseline model, we use \textbf{word-based} tokenization, which uses whitespace as a separator between words in a caption. We filter out the words that appear less than 5 times in the training set, resulting into a vocabulary of 9,486 words/tokens.
 
 For our model, we use SentencePiece \cite{kudo2018sentencepiece}, which is a language independent \textbf{subword} tokenization and detokenzation tool. SentencePiece works on raw data and does not require any initial tokenization of words. It employs various efficient techniques for training and segmentation with raw data, to perform subword tokenization. The raw input text is treated as a sequence of Unicode characters and even the whitespace is treated as a normal symbol `$\_$' (U+2581). 

Using SentencePiece, we specify Unigram language model as our choice of subword segmentation algorithm and  also set out the required vocabulary size. We experiment with different vocabulary sizes and analyze the difference in performance, the details of which are in Section~\ref{subsection:Influence of the Vocabulary Size}. Moreover, to train the baseline and our proposed models we truncate all the  captions in the training set which are longer than 16 words.
\begin{table*}[t]
	\tiny
	\centering
	\resizebox{\textwidth}{!}{%
		\begin{tabular}{@{}|l|c|c|c|c|c|c|c|c|c|c|@{}}
			\toprule
			\toprule
			Models   & \begin{tabular}[c]{@{}c@{}}Vocab \\ Size\end{tabular} & \begin{tabular}[c]{@{}c@{}}Model\\ Parameters\end{tabular} & B1            & B2            & B3            & B4            & M             & R            & C              & S             \\ \midrule
			Baseline & \textbf{9,486}                                                 & \textbf{54.9M}                                                      & 75.2          & 58.8          & 44.6          & 33.7          & 27.5          & 55.5          & 111.0          & 21.0          \\
			SubICap-3k   & 3,078                                                 & 48.3M                                                      & 75.3          & 59.1          & 45.1          & 34.3          & 28.1         & 55.5          & 107.0          & 19.7          \\
			SubICap-2k   & 2,079                                                 & 47.3M                                                      & 76.7          & 60.8          & 47.0          & 36.1          & 29.4          & 56.7          & 114.6          & 20.9           \\
			SubICap-1k   & 1,085                                                 & 46.3M                                                      & \textbf{76.7} & \textbf{60.8} & \textbf{47.1} & \textbf{36.2} & \textbf{29.7} & \textbf{56.9} & \textbf{116.1} & \textbf{21.1}   \\
			SubICap-500  & 579                                                   & 45.8M                                                      & 75.8          & 59.7          & 46.1          & 35.4          & 29.5          & 56.4          & 114.0          & 21.0            \\
			SubICap-300  & 335                                                   & 45.5M                                                      & 75.9          & 59.7          & 45.9          & 35.2          & 29.3          & 56.1          & 113.7          & 21.0           \\ \bottomrule
	\end{tabular}}
	\caption{Shows the comparison between, training vocabulary size, trainable model parameters, and metric scores of models on the MSCOCO offline test set. All the models are trained for the standard cross-entropy loss for 30 epochs and use a beam size of 2 for inference. Highest values are shown in boldface}
	\vspace{-2em}
	\label{tab:comparison}
\end{table*}

\section{Dataset and Metrics}
For all our experiments we use one of the most widely used public captioning dataset MSCOCO \cite{chen2015microsoft} which contains 123,287 images, each paired with atleast 5 captions in English language. To maintain consistency with literature, we use the publicly available split of MSCOCO\footnote{https://cs.stanford.edu/people/karpathy/deepimagesent/} which provides 5,000 images for validation and testing each. We report our results on the offline MSCOCO test set, comprising of 5000 images.  
We evaluate the performance of captioning models, using the commonly used metrics such as BLEU \cite{papineni2002bleu}, METEOR \cite{banerjee2005meteor}, ROUGE-L \cite{lin2004rouge}, CIDEr-D \cite{vedantam2015cider} and SPICE \cite{anderson2016spice}. We obtain the scores of these metric using MSCOCO evaluation toolkit \footnote {https://github.com/tylin/coco-caption}. For the sake of brevity, we re-label the metric names BLEU1,  BLEU2, BLEU3, BLEU4, METEOR, ROUGE-L, CIDEr-D and SPICE to B1, B2, B3, B4, M, R, C, and S respectively in our Tables.   
\section{Experiments and Results}

\begin{figure}
	\centering
	\includegraphics[width=7cm]{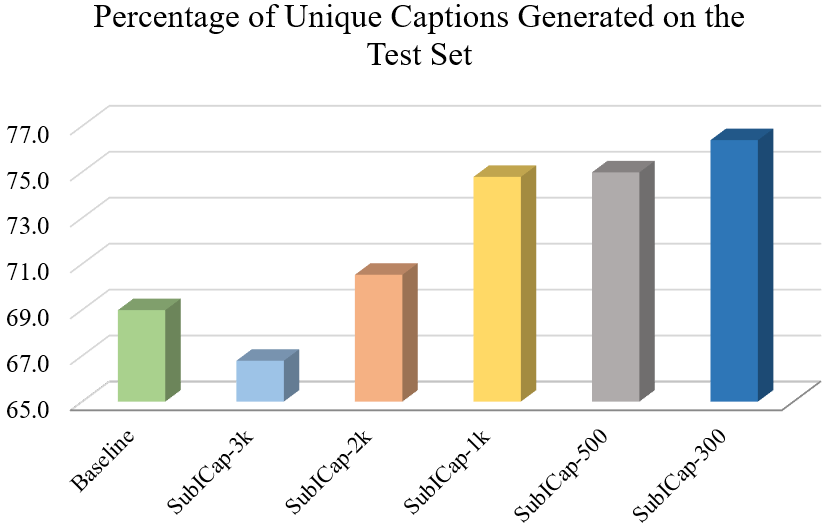}
	\caption{Comparison between the percentage of unique captions generated on the MSCOCO offline test set.}
	\label{fig:vocab}
	\vspace{-1.5em}
\end{figure}

\subsection{Influence of the Vocabulary Size} 
\label{subsection:Influence of the Vocabulary Size}
Unigram language model based tokenizer requires a predefined vocabulary size of tokens/subwords, to perform unsupervised segmentation. As explained in Sec.~\ref{subsection:Subword Tokenization}, once the desirable vocabulary size is defined, the unigram model then seeks to iteratively create a vocabulary of subwords. 

Therefore, the vocabulary size is an important parameter and can impact the learning of a language model. We seek to analyse the impact of various vocabulary sizes on the performance of our captioning model. An optimal balance is searched between the compactness of the subword vocabulary and that of caption representation. Generally, a lower subword vocabulary size leads to a less compact caption representation and vice versa.  As specifying a smaller vocabulary enforces the unigram model to represent all the words in the corpus with fewer subwords, it has to breakdown words into smaller subwords, and the resulting captions are longer sequences of subwords. Following is an arbitrary example of token sequences generated by the unigram language model tokenizer based on different vocabulary sizes. 

\begin{itemize}[noitemsep,topsep=0pt]
	\item \textbf{Raw Caption}: a cat is climbing a tree
	\item \textbf{ULM-1k}: [a][cat][is][climb][\_ing][a][tree]
	\item \textbf{ULM-300}: [a][cat][is][c][\_l][\_i][\_m][\_b][\_ing][a][tree]
\end{itemize}

\noindent where, ULM-1k, and ULM-300 are the names of the Unigram language models based on 1k, and 300 vocabulary size, respectively. Various variants of our model are named SubICap-3k, SubICap-2k, SubICap-1k, SubICap-500, SubICap-300, where the suffix represents the vocabulary size. For example SubICap-1k uses a vocabulary of 1k subwords.

Table~\ref{tab:comparison} shows the comparison of vocabulary size, trainable parameters and metric scores of our models, trained on MSCOCO dataset for 30 epochs using cross-entropy loss. It can be seen that various variants of our model outperform the word-level baseline in terms of the metric scores, where, SubICap-1k shows the best performance amongst all. As discussed in literature, subwords generated by learned segmentation algorithms often resemble linguistic morphemes \cite{ creutz2006induction}. The subword vocabulary size tends to impact the quality of morphemes/subwords generated by the Unigram model and that is one of the reason why we observe a difference in performance due to the vocabulary size. Moreover, to our understanding SubICap-1k segments the words in a better way compared to other variants.

We also compare the percentage of unique captions generated by the baseline and our proposed model, shown in Figure ~\ref{fig:vocab}. Note that the training vocabulary sizes of our proposed models are significantly smaller than those of the baseline, however, the percentage of unique captions generated by our models (except for SubICap-3k) is greater than that of the baseline.

\begin{table}[t]
	\small
	\begin{center}
		\begin{tabular}{@{}|l|c|c|c|@{}}
			\toprule
			Models     &  \begin{tabular}[c]{@{}c@{}}Vocab. \\ Size\end{tabular} & \begin{tabular}[c]{@{}c@{}}Unique\\ Captions (\%age)\end{tabular} & \begin{tabular}[c]{@{}c@{}}Avg. Caption \\ Length (words)\end{tabular} \\ \midrule
			Baseline   & 9,486      & 83.5                                                               & 9.4                                                                    \\
			SubICap-1k & 1,085      & \textbf{89.0}                                                      & \textbf{10.1}                                                          \\ \bottomrule
		\end{tabular}%
	\end{center}
	\vspace{-0.5em}
	\caption{Comparison between our model (SubICap) and baseline in terms of percentage of unique captions and descriptiveness after optimizing both for CIDEr-D score for 10 epochs.}
	\vspace{-1.5em}
	\label{tab:unique table}
\end{table}


\begin{table*}[]
\tiny
\centering
\resizebox{\textwidth}{!}{%
\begin{tabular}{@{}llcccccccccccc@{}}
\toprule
                  &                      & \multicolumn{6}{|c}{\textbf{Cross-Entropy Loss}}                                                                                                       & \multicolumn{6}{|c|}{\textbf{CIDEr-D Optimization}}                                                                   \\ \midrule

\multicolumn{1}{|l|}{Models}            & \multicolumn{1}{l|}{Vocab size} & B1            & B4            & M             & R             & C              & \multicolumn{1}{c|}{S}             & B1            & B4            & M             & R             & C              & \multicolumn{1}{c|}{S}             \\ \midrule
\multicolumn{1}{|l|}{ReviewNet \cite{yang2016review}}        & \multicolumn{1}{l|}{9,520}      & -             & 29.0          & 23.7          & -             & 88.6           & \multicolumn{1}{c|}{-}             & -             & -             & -             & -             & -              & \multicolumn{1}{c|}{-}             \\
\multicolumn{1}{|l|}{ACVT \cite{wu2016value}}              & \multicolumn{1}{l|}{8,791}      & 74.0          & 31.0          & 26.0          & -             & 94.0           & \multicolumn{1}{c|}{-}             & -             & -             & -             & -             & -              & \multicolumn{1}{c|}{-}             \\
\multicolumn{1}{|l|}{COMIC \cite{tan2019comic}  }             & \multicolumn{1}{l|}{258}        & 72.9          & 32.8          & -             & -             & 100.1          & \multicolumn{1}{c|}{18.5}          & 75.3          & 34.4          & -             & -             & 105.0          & \multicolumn{1}{c|}{19.0}          \\
\multicolumn{1}{|l|}{UpDown \cite{anderson2017bottom}}            & \multicolumn{1}{l|}{10,010}     & 77.2          & 36.2          & 27.0          & 56.4          & 113.5          & \multicolumn{1}{c|}{20.3}          & 79.8          & 36.3          & 27.7          & 56.9          & 120.1          & \multicolumn{1}{c|}{21.4}          \\
\multicolumn{1}{|l|}{GCN-LSTM \cite{yao2018exploring}}          & \multicolumn{1}{l|}{10,201}     & 77.3          & 36.8          & 27.9          & {57.0}          & {116.3}          & \multicolumn{1}{c|}{20.9}          & \textbf{80.5}          & 38.2          & 28.5          & {58.3}          & {127.6}          & \multicolumn{1}{c|}{22.0}          \\
\multicolumn{1}{|l|}{RFNet \cite{jiang2018recurrent}}             & \multicolumn{1}{l|}{9,487}      & 76.4          & 35.8          & 27.4          & 56.8          & 112.5          & \multicolumn{1}{c|}{20.5}          & 79.1          & 36.5          & 27.7          & 57.3          & 121.9          & \multicolumn{1}{c|}{21.2}          \\
\multicolumn{1}{|l|}{AoANet \cite{huang2019attention}}            & \multicolumn{1}{l|}{10,369}     & \textbf{77.4} & \textbf{37.2} & 28.4         & \textbf{57.5} & \textbf{119.8} & \multicolumn{1}{c|}{\textbf{21.3}} & 80.2          & \textbf{38.9} & {29.2}          & \textbf{58.8} & \textbf{129.8} & \multicolumn{1}{c|}{22.4}         

\\ \midrule
\multicolumn{1}{|l|}{Baseline* \cite{herdade2019image}}          & \multicolumn{1}{l|}{9,486}      & 75.2          & 33.7          & 27.5          & 55.5          & 111.0          & \multicolumn{1}{c|}{21.0}          & 80.2          & 38.1          & 28.7          & 58.2          & 127.3          & \multicolumn{1}{c|}{22.3}          \\

\multicolumn{1}{|l|}{SubICap-1k} & \multicolumn{1}{l|}{1,085}      & 76.7          & 36.2          & \textbf{29.7} & 56.9          & 116.1          & \multicolumn{1}{c|}{21.1}          & 79.5          & 37.1          & \textbf{29.8} & 58.2          & 123.2          & \multicolumn{1}{c|}{\textbf{23.0}} \\ \bottomrule
\end{tabular}%
}
\vspace{0.5em}
\caption{Comparison against state-of-the-art models trained for cross-entropy loss and fine-tuned using self-critical training, on MSCOCO offline test set. To maintain consistency with literature \cite{herdade2019image}, we generate captions for models (baseline and SubICap) trained with cross-entropy loss and CIDEr-D optimization, setting beam size to 2 and 5, respectively. * indicates the results obtained from a publicly available model. Highest scores are shown in bold face.}
\vspace{-2em}
\label{tab:rl table}
\end{table*}

Based on our comparison, we choose our model SubICap-1k for further experiments as it strikes a balance between the vocabulary size and compactness of caption representation. Moreover, it also achieves the highest metric scores compared to the other models.

\subsection{Caption Uniqueness and Descriptiveness}
\label{subsection:Caption Uniqueness and Descriptiveness}
Uniqueness is an important aspect of caption quality. Humans tend to be canny in terms of generating diverse captions, and avoid repeating same generic captions. Existing neural models suffer from the problem of regurgitating captions from the training set. The output captions are usually generic, which can safely attain a higher metric score, but are less discursive \cite{herdade2019image}. Table~\ref{tab:unique table} shows that our model generates a higher percentage of unique captions (89.0\%) compared to the baseline (83.5\%). Moreover, the captions generated by our model are more descriptive, which is reflected by the higher average length (no. of words). This demonstrates the usefulness of subword modeling for the caption generation task, which not only helps improve the quantitative scores but also the uniqueness and descriptiveness of captions.

\section{Comparison with State-of-the-art Models}

\subsection{Quantitative Analysis}
\label{subsection:Quantitative Analysis}
Table~\ref{tab:rl table} shows the comparison between the metric scores and training vocabulary size of our model against the SOTA models. First and foremost, it can be observed from Table~\ref{tab:rl table} that all the SOTA models, except for \cite{tan2019comic} have been trained on a vocabulary 8 to 10 times greater than our model. COMIC \cite{tan2019comic} does use the smallest vocabulary, however, it significantly lags behind in terms of the metrics' scores. In the case of cross-entropy loss based training, SubICap-1k not only improves upon the baseline, but it also shows a competitive performance to AoANet (best performing model), which uses 10 times larger vocabulary than our model. We believe that integrating the subword-information to any SOTA model can improve its performance as we have demonstrated for the baseline. 

We further optimize our model SubICap-1k and the baseline for CIDEr-D score \cite{rennie2017self} using self-critical training. Table~\ref{tab:rl table} also shows the comparison of metric scores of our model against the SOTA, after self-critical training. We notice an improvement in METEOR and SPICE scores (2.01\% and 3.13\% respectively) of SubICap-1k compared to the baseline. SubICap-1k also achieves the highest SPICE and METEOR scores amongst the SOTA models. This shows that our model generates captions which are semantically better than those generated by other models \cite{sharif2019lceval}. 

From Table~\ref{tab:rl table}, we also observe that our model achieves a lower CIDEr score (3.2\%) compared to the the baseline. CIDEr is an n-gram based metric, which predominantly captures the lexical/syntactic correspondence between the generated and reference captions. As mentioned in the literature \cite{anderson2016spice}, n-gram based metric scores might not always be the best reflection of caption quality \cite{anderson2016spice}. It is quite possible for two captions to differ in terms of words or structure, but carry the same meaning. Another reason behind our model's lower CIDEr score compared to the baseline can be that the baseline tends to regurgitate a higher percentage of the training captions (Table~\ref{tab:rl table}), which are syntactically closer to the ground truth (since they are written by humans), and thus achieve a higher CIDEr score. 

SPICE which captures the semantic quality of captions \cite{anderson2016spice}, shows a comparatively higher correlation with human scores in terms of captioning model assessment \cite{anderson2016spice}. Higher SPICE and METEOR scores represent better semantic as well as lexical quality of captions. We also share some qualitative examples in Sec. 7.2 to clarify our point of view.

\subsection{Qualitative Analysis}
\label{subsection:Qualitative Analysis}

\begin{figure}
	\centering
	\includegraphics[width=8.25cm]{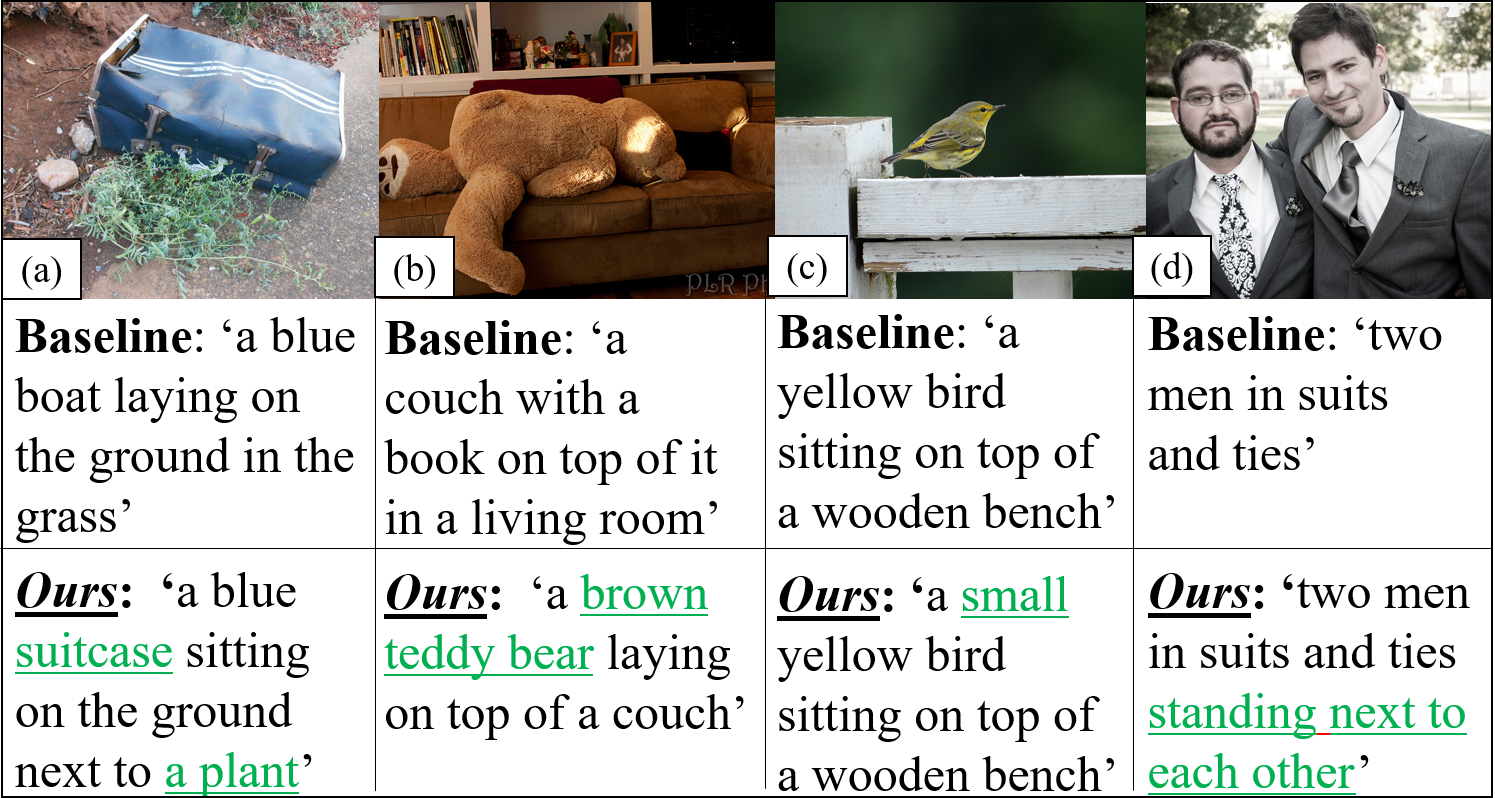}
	\caption{Shows the examples of captions and images for which our model (SubICap) achieves higher \textbf{SPICE} score. Our model performs better in terms of objects (a), color and object (b), size (c), and relationship (d). Improvements are shown in green color.}
	\vspace{-1em}
	\label{fig:spice_qual2}
\end{figure}

\begin{figure}
	\centering
	\includegraphics[width=8.25cm]{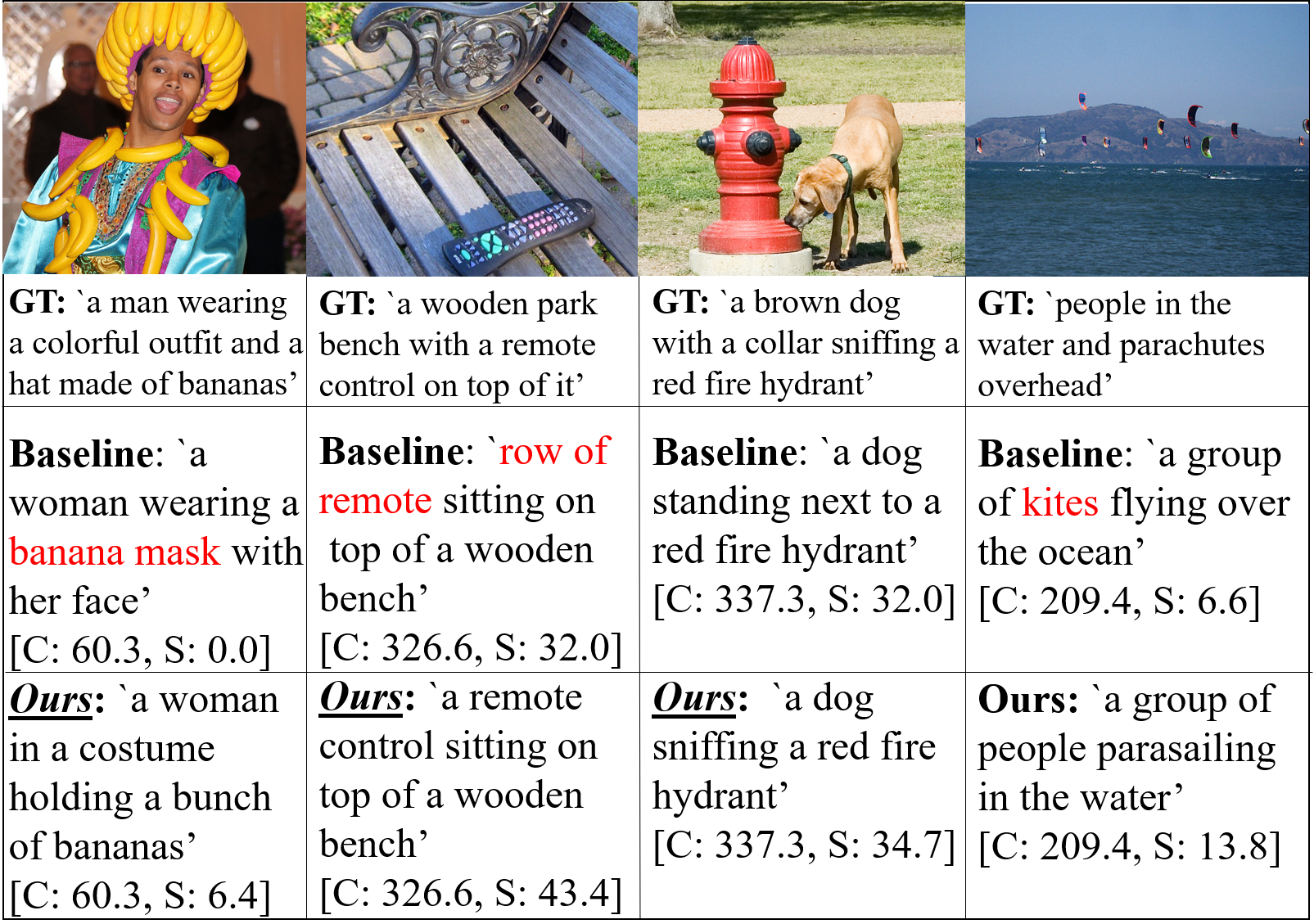}
	\caption{A comparison of captions generated by our model (SubICap-1k) vs. baseline, which are optimized for CIDER-D. Scores of CIDEr-D (C) and SPICE (S) are provided, along with the generated and ground truth captions. Mistakes are shown in red color.}
	\vspace{-1em}
	\label{fig:cider}
\end{figure}

While it is a standard practice to compare models in terms of metric scores, it is also important to have a qualitative analysis, because the commonly used measures have various limitations \cite{kilickaya2016re}. Figure~\ref{fig:spice_qual2} and Figure~\ref{fig:cider} show various examples of captions generated by our model and the baseline for qualitative comparison. As our model achieved the SOTA SPICE scores on MSCOCO offline test split, we provide a further breakdown of SPICE metric in Table~\ref{tab:spice-comp}. It can be observed from Table~\ref{tab:spice-comp}, that our metric improves upon the color, attribute, object, relation and size scores of the baseline. In Figure~\ref{fig:spice_qual2}, we provide examples of captions generated by our model which achieve a higher SPICE score compared to the ones generated by the baseline.

\begin{table}[t]
	\small
	\begin{center}
		\begin{tabular}{@{}|l|l|l|l|l|l|l|@{}}
			\toprule
			\multicolumn{1}{c}{}           & Color        & Att.     & Obj.        & Rel.     & Card.  & Size         \\
			\toprule
			\toprule
			Baseline              & 5.8 & 10.9          & 40.5          & 6.6          & \textbf{5.0}          & 1.2          \\
			SubICap-1k                & \textbf{6.2}          & \textbf{12.1} & \textbf{40.9} & \textbf{7.0} & 4.6 & \textbf{1.9}
			
			\\ \bottomrule	
		\end{tabular}%
	\end{center}
	\vspace{-1em}
	\caption{Shows the breakdown of SPICE metric scores of our model and baseline for various categories (Color, Attributes, Objects, Relationships, Cardinality and Size). The scores reported are of fine-tuned models using self-critical training for 10 epochs.}
	\vspace{-0.2em}
	\label{tab:spice-comp}
\end{table}

As discussed earlier in Sec. 7.1, the $n$-gram based measures tend to overlook the semantics and only focus on the lexical properties of the captions. CIDEr, which is an $n$-gram based measure, \cite{sharif2019lceval} prefers captions which have a higher lexical correspondence to the ground truth caption. However, in various cases, it is quite possible that two captions which have different words or structure, might carry the same meaning and vice versa. To present our point of view, we show a few examples in Figure~\ref{fig:cider}, where CIDEr gives an equal or a lower score to the caption generated by our model over the baseline's. It can be observed that the captions generated by our model, better reflect the visual content and are semantically closer to the ground truth caption, thus they gain a higher SPICE score. Moreover, the captions generated by our model are also lexically sound (high METEOR score) and more descriptive (Fig.~\ref{fig:spice_qual2}, Table~\ref{tab:unique table}). Overall, our model shows a promising qualitative and quantitative performance.

\subsection{Conclusion}

In this work we proposed a subword-informed captioning model, which treats captions as a sequence of subword units. In contrast to the existing IC models, which do not leverage upon the semantic composition of words, and ignore rare words, our model exploits the relationship between word segments and effectively handles rare words. Our experimental results show that our model, not only reduces the trainable model parameters, but also significantly improves upon the quality of captions in terms of uniqueness and descriptiveness. Moreover, our model achieves state-of-the-art (SOTA) performance in terms of SPICE and METEOR scores. SPICE and METEOR  are the two metrics which correlate higher with human judgments compared to the other commonly used ones  \cite{anderson2016spice}. In future, we plan to investigate the impact of subword tokenization for image captioning models developed for other languages such as Chinese and Japanese. 

\section*{Acknowledgments}
This work is supported by Australian Research Council, ARC DP150100294. We are grateful to NVIDIA for providing Titan-Xp GPU, which was used for the experiments.

\newpage
\newpage
{\small
\bibliographystyle{ieee_fullname}
\bibliography{egbib}
}

\end{document}